\documentclass[runningheads]{llncs}

\usepackage{eccv}
\usepackage[accsupp]{axessibility}
\usepackage[breaklinks,colorlinks,citecolor=eccvblue]{hyperref}
\usepackage{eccvabbrv}
\usepackage{graphicx}
\usepackage{amsmath}
\usepackage{array}
\usepackage{pifont}
\usepackage{placeins}
\usepackage{float} % in preamble
\usepackage{graphicx}
\usepackage{booktabs}
\usepackage{amssymb}
\usepackage[accsupp]{axessibility}
\newcommand{\method}{DisasterInsight}
\newcommand{\cmark}{\ding{51}}  % ✓
\newcommand{\xmark}{\ding{55}}  % ✗

\usepackage[most]{tcolorbox}
\usepackage{multirow}
\usepackage{orcidlink}

\usepackage{multicol}
\begin{document}

\title{DisasterInsight: A Building-Centric Benchmark for Evaluating Vision--Language Models in Disaster Response}
\titlerunning{DisasterInsight}
% For review version, keep anonymous:
\author{Sara Tehrani\inst{1}\orcidlink{0009-0004-6811-9742} \and
Yonghao Xu\inst{1}\thanks{Corresponding author.}\orcidlink{0000-0002-6857-0152} \and
Leif Haglund\inst{1,2}\orcidlink{0009-0006-0801-2024} \and
Amanda Berg\inst{1,2}\orcidlink{0000-0002-6591-9400} \and \\ 
Gulnaz Zhambulova\inst{1}\orcidlink{0009-0008-0665-0393} \and
Michael Felsberg\inst{1}\orcidlink{0000-0002-6096-3648}  
}
\authorrunning{S. Tehrani et al.}
\institute{
Computer Vision and Learning System, Linköping University, Sweden
\and
Vantor, Linköping, Sweden
}

\maketitle
\begin{abstract}
Vision--language models (VLMs) show promise for disaster-response remote sensing, but existing benchmarks mainly emphasize scene-level or damage-centric assessment.
To study this building-centric gap, we introduce \method{}, a diagnostic benchmark built on xBD, a pre/post-disaster satellite dataset with building-level damage labels.
\method{} enriches building instances with OpenStreetMap-derived functional labels and contains 134{,}108 task-specific instruction records across 15 task types, spanning instance-level assessment, scene-level counting, multi-instance reasoning, and structured report generation.
The benchmark supports RGB pre/post-disaster imagery, single- and multi-view instance formulations, and scene-level RGB/SAR diagnostic inputs.
Experiments with general-domain and remote-sensing VLMs show that models perform better on visible damage cues than on building-function understanding, multi-instance reasoning, counting, and grounded reporting.
Instruction tuning improves performance on several tasks but does not close this building-centric gap. The code and benchmark resources are available at \href{https://github.com/sara2227/DisasterInsight-benchmark}{DisasterInsight benchmark repository}.
\end{abstract}

\section{Introduction}
\label{sec}

Rapid and accurate interpretation of satellite imagery is a cornerstone of effective disaster response.
Following large-scale hazards such as earthquakes, floods, or hurricanes, emergency coordination agencies require fine-grained, building-level assessments to support triage, resource allocation, and risk mitigation decisions~\cite{dong2013comprehensive}.
Operational disaster-response guidelines from the International Search and Rescue Advisory Group (INSARAG)\footnote{\scriptsize{INSARAG Guidelines, Vol.~II, Manual~B, Sec.~5.8.2, p.~37, available at \url{https://insarag.org/wp-content/uploads/2021/06/INSARAG20Guidelines20Vol20II2C20Man20B.pdf}, accessed Sep.~2026.}} and the Federal Emergency Management Agency (FEMA)\footnote{\scriptsize{FEMA Preliminary Damage Assessment Guide, p.~21, available at \url{https://www.fema.gov/sites/default/files/2020-07/fema_preliminary-disaster-assessment_guide.pdf}, accessed Sep.~2026.}} further emphasize that responders must assess not only physical damage, but also the function and use of affected structures.
A damaged hospital, school, shelter, industrial facility, or residential building may imply different priorities for medical surge capacity, shelter availability, search-and-rescue planning, and infrastructure restoration.

Despite recent advances in vision--language models (VLMs), existing remote-sensing benchmarks do not adequately capture these operational requirements.
Most datasets emphasize coarse damage labels or scene-level analysis, treating buildings as an undifferentiated class and overlooking functional semantics and per-building reasoning~\cite{chen2025bright,rahnemoonfar2021floodnet, gupta2019creating}.
As illustrated in Figure~\ref{fig}, a scene-level summary may describe aggregate disaster severity while missing which specific buildings are affected, what functions they serve, and why they matter for response prioritization.
Recent multimodal benchmarks incorporating natural-language outputs have explored captioning and question answering over disaster imagery, but typically rely on unstructured scene-level narratives that are not explicitly grounded to individual buildings or aligned with humanitarian reporting practices~\cite{irvin2025teochat,wang2026disasterm3,kuckreja2024geochat}.

To address these limitations, we introduce \method{}, a diagnostic benchmark for building-centric disaster understanding. Built on xBD~\cite{gupta2019creating}, \method{} enriches building instances with OpenStreetMap-derived functional labels and supports instance-level building assessment, scene-level reasoning, multi-instance reasoning, and structured report generation across single-view, multi-view, RGB, and scene-level RGB/SAR settings. The benchmark is designed to expose where current VLMs succeed on visible damage cues and where they remain unreliable in building-function understanding, multi-building reasoning, and grounded disaster reporting.

Our experiments reveal a consistent limitation: VLMs are stronger on visible damage cues than on building-function recognition, multi-instance reasoning, scene-level counting, and grounded report generation.
Instruction tuning improves damage and temporal-change reasoning, but building-function recognition remains limited, especially under the long-tailed OSM-derived category distribution.
These results suggest that current VLMs still struggle to ground higher-level semantic reasoning in building-level visual evidence.

\textbf{Contributions.}
\begin{itemize}
\item We introduce \method{}, an instance-centric disaster-response VLM benchmark that enriches xBD buildings with OSM-derived functional labels, enabling evaluation of building identity, function, damage, and response relevance.

\item We extend xBD with co-registered Sentinel-1 SAR data and use it in a subset of scene-level RGB/SAR diagnostic inputs, enabling analysis of complementary optical--radar evidence for disaster understanding.

\item We construct 134{,}108 instruction records across 15 task types and evaluate general-domain and remote-sensing VLMs under a unified protocol with instruction tuning. Results show that tuning improves damage and temporal-change reasoning, but models still struggle with building function, multi-instance reasoning, counting, and grounded reporting.
\end{itemize}

\begin{figure}[t]
\centering
\includegraphics[width=0.7\textwidth]{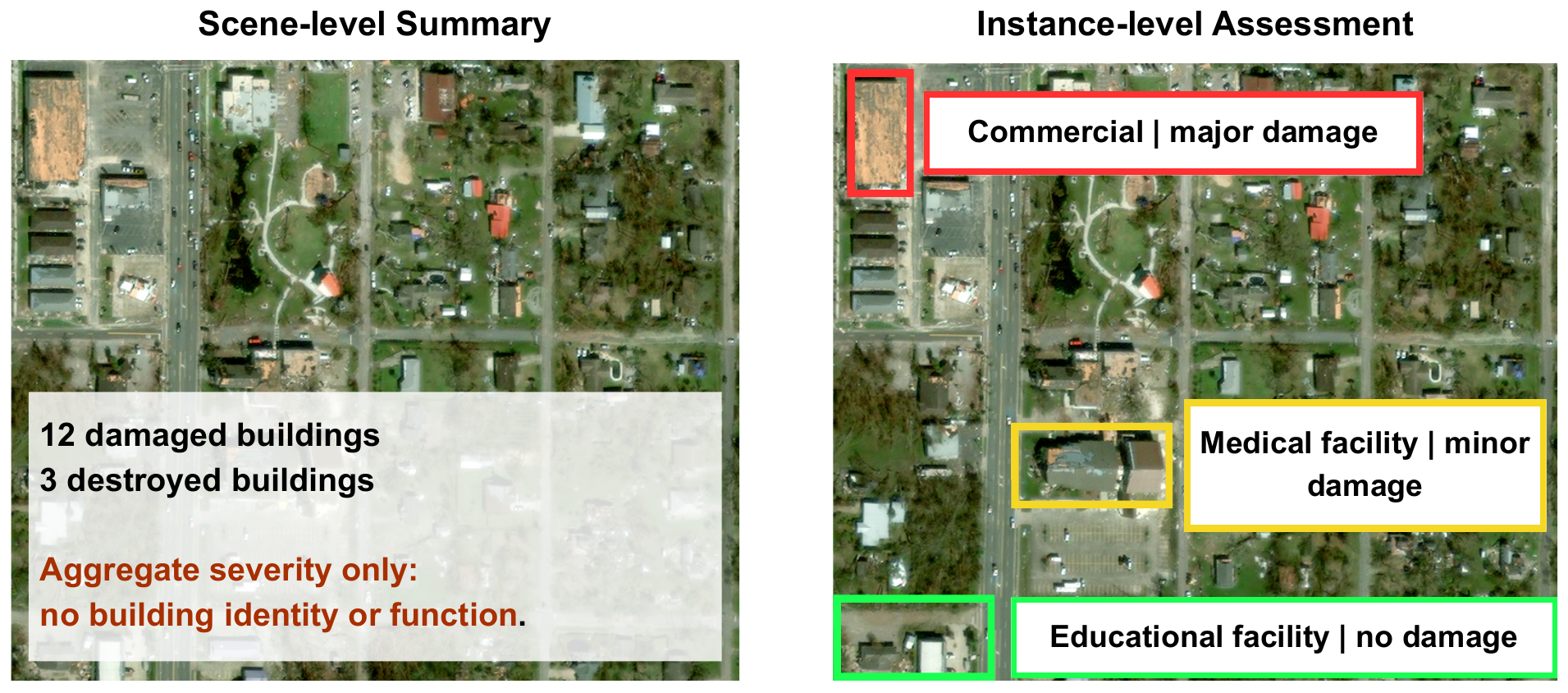}
\caption{
Scene-level summaries can miss building-level evidence.
\method{} grounds disaster assessment at the building-instance level.
}
\label{fig}
\end{figure}

\section{Related Work}
\label{sec:related_work}

\subsection{Disaster Assessment Datasets}
\label{subsec:disaster_datasets}
Remote sensing (RS) datasets have enabled substantial progress in disaster assessment through damage mapping, flood analysis, and change detection.
Datasets such as xBD~\cite{gupta2019creating}, BRIGHT~\cite{chen2025bright}, and FloodNet~\cite{rahnemoonfar2021floodnet} provide pre- and post-event imagery with disaster-related annotations across multiple hazards.
However, these datasets are primarily designed for damage-centric or scene-level assessment.
Even when building footprints and damage labels are available, buildings are typically treated as damage-bearing objects without explicit information about their functional role.
As a result, existing datasets do not evaluate whether models can distinguish, for example, residential, medical, educational, commercial, or emergency-related buildings, which is essential for response prioritization in real-world disaster scenarios.
\subsection{Vision--Language Models in Remote Sensing}
\label{subsec:vlms_rs}
Recent work extends vision--language models (VLMs) to remote sensing by adapting instruction-following and LLaVA-style architectures to satellite imagery, including GeoChat~\cite{kuckreja2024geochat}, TEOChat~\cite{irvin2025teochat}, and EarthDial~\cite{soni2025earthdial}.
Generic VLMs such as LLaVA-OneVision~\cite{li2024llavaonevision} and Qwen-VL~\cite{bai2023qwen} have also been explored for remote-sensing interpretation through zero-shot or instruction-following evaluation.
These models provide a natural interface for querying complex scenes and generating textual outputs, making them attractive for disaster-response applications.
Nevertheless, existing RS VLMs and benchmarks mainly evaluate broad scene understanding, captioning, visual question answering, or coarse grounding.
They provide limited diagnosis of fine-grained, per-building semantics, especially when models must reason about building function, damage state, and multi-building relationships under disaster conditions.

\subsection{Building-Centric and Grounded Disaster Reasoning}
\label{subsec:building_grounded_reasoning}

Grounded disaster reasoning requires models to connect visual evidence to specific structures and produce outputs useful for operational decision-making.
Prior crisis summarization and event-analysis work uses social media, textual reports, or other non-visual sources for situational awareness~\cite{imran2020using,purohit2014whom,vieweg2010microblogging}, but generally lacks visual grounding to individual buildings.
Recent multimodal disaster benchmarks such as DisasterM3~\cite{wang2026disasterm3} move toward vision--language evaluation for disaster assessment, including perception, reasoning, and report generation.
However, they do not explicitly evaluate building-function understanding or jointly test building identity, function, damage, and multi-instance reasoning.

Table~\ref{tab:comparison} summarizes these differences.
Existing datasets cover parts of the problem, such as damage assessment, visual grounding, multimodal inputs, or reporting, but none jointly provide instance-level grounding, building-function labels, damage assessment, multi-instance reasoning, and structured reports.
\method{} addresses this gap by enriching xBD building instances with OSM-derived functional labels and evaluating VLMs across instance-level, scene-level, multi-instance, and report-generation tasks.

\begin{table*}[tb]
\centering
\scriptsize
\caption{
Comparison with representative disaster and vision--language datasets.
``Bldg. Inst.'' indicates building-instance grounding, and ``Function'' indicates explicit building-use labels.% and RGB imagery is treated as optical imagery.
}
\label{tab:comparison}
\renewcommand{\arraystretch}{1.12}
\setlength{\tabcolsep}{3pt}
\resizebox{\textwidth}{!}{%
\begin{tabular}{lccccc}
\toprule
\textbf{Dataset}
& \textbf{Bldg. Inst.}
& \textbf{Damage}
& \textbf{Modality}
& \textbf{Report}
& \textbf{Function} \\
\midrule
RSICD~\cite{lu2017exploring}
& \xmark & \xmark & Optical + caption & Caption & \xmark \\

BRIGHT~\cite{chen2025bright}
& \xmark & \cmark & Optical/SAR & \xmark & \xmark \\

FloodNet~\cite{rahnemoonfar2021floodnet}
& \xmark & \cmark & UAV RGB + QA & VQA & \xmark \\

xBD~\cite{gupta2019creating}
& \cmark & \cmark & Pre/post optical & \xmark & \xmark \\

CrisisMMD~\cite{alam2018crisismmd}
& \xmark & Partial & Image + text & Summ. & \xmark \\

CrisisFACTS~\cite{mccreadie2023crisisfacts}
& \xmark & \xmark & Text & Summ. & \xmark \\

DisasterQA~\cite{rawat2024disasterqa}
& \xmark & \xmark & Text & QA & \xmark \\

GeoChat~\cite{kuckreja2024geochat}
& \xmark & \xmark & Optical + instr. & \xmark & \xmark \\

TEOChat~\cite{irvin2025teochat}
& \cmark & \cmark & Optical + instr. & Caption & \xmark \\

DisasterM3~\cite{wang2026disasterm3}
& \cmark & \cmark & Optical/SAR + instr. & \cmark & \xmark \\

\midrule
\textbf{DisasterInsight (Ours)}
& \textbf{\cmark}
& \textbf{\cmark}
& \textbf{Optical/SAR + instr.}
& \textbf{\cmark}
& \textbf{\cmark} \\
\bottomrule
\end{tabular}%
}
\end{table*}
\section{DisasterInsight Benchmark}
\label{sec:benchmark}
\subsection{Benchmark Motivation: From Scene-Level to Building-Centric Evaluation}
\label{subsec:motivation}
Most disaster-response benchmarks evaluate whether a model can recognize that a scene is damaged, flooded, burned, or otherwise affected.
While useful, this scene-level perspective is often insufficient for operational decision-making.
A single post-disaster image may contain many buildings with different damage states, functional roles, and response priorities.
For example, a damaged medical facility, school, shelter, or industrial building may require a different response than a damaged residential structure.
Disaster understanding should therefore address not only what happened in the scene, but also which buildings are affected, what functions they serve, and how multiple structures should be compared or prioritized.

\begin{figure}[t]
\centering
\includegraphics[width=0.95\textwidth]{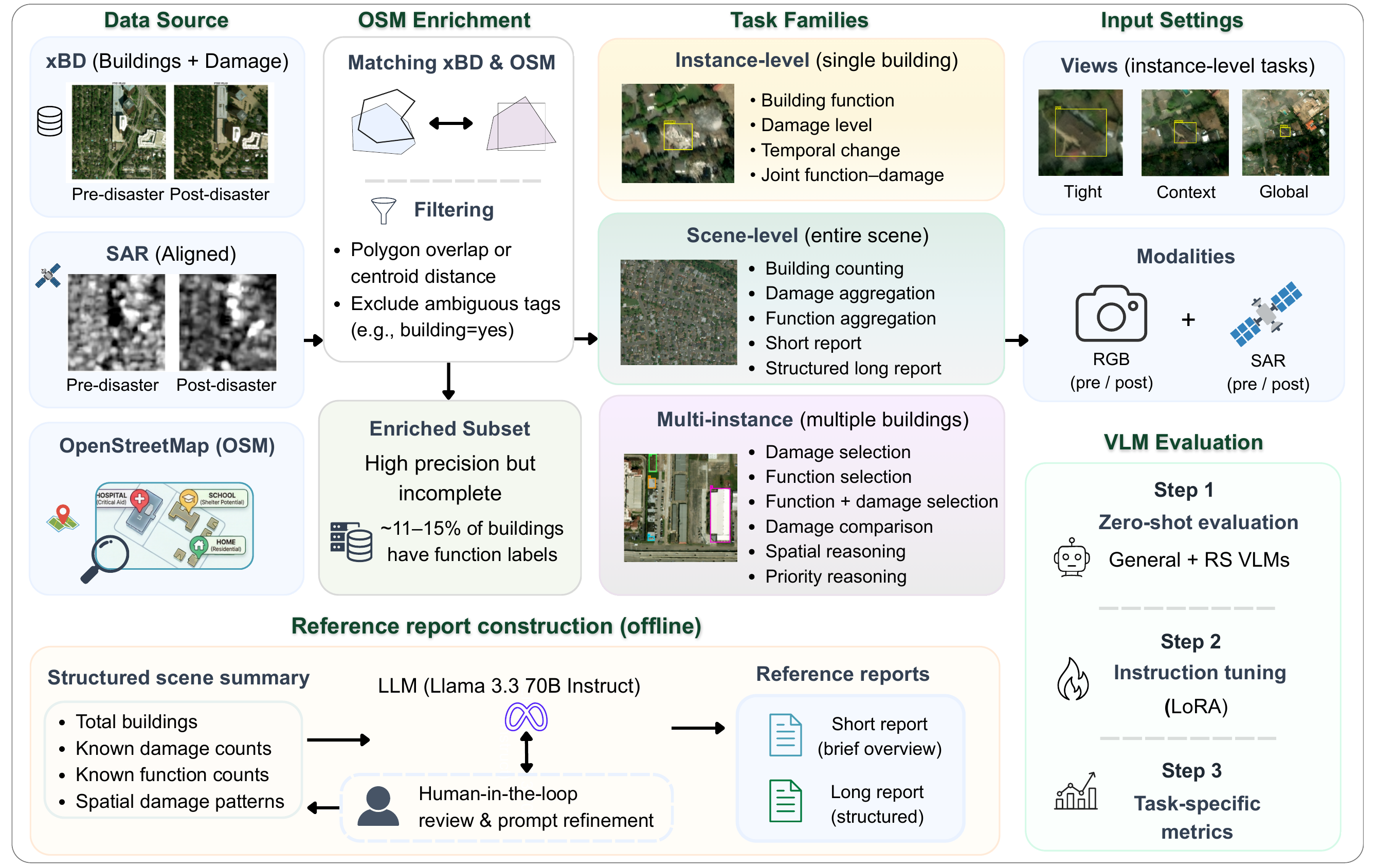}
\caption{Overview of \method{}: OSM-enriched xBD buildings and tasks for instance, scene, multi-instance, and report generation. © ESA/Sentinel.}
\label{fig:benchmark_overview}
\end{figure}

\method{} is designed to evaluate this building-centric form of reasoning.
As shown in Figure~\ref{fig:benchmark_overview}, \method{} enriches xBD pre/post imagery and building footprints with OSM-derived functional labels, then converts them into instruction-following tasks for instance-level, scene-level, multi-instance, and report-generation evaluation across RGB and RGB/SAR settings.
% As shown in Figure~\ref{fig:benchmark_overview}, the benchmark starts from xBD pre- and post-disaster imagery and building footprints, enriches building instances with OSM-derived functional labels, and converts the resulting annotations into instruction-following tasks.
% These tasks cover instance-level assessment, scene-level reasoning, multi-instance reasoning, and structured report generation under single-view, multi-view, grounded, and scene-level optical/SAR diagnostic input settings.

\subsection{Data Sources and Instance Construction}
\label{subsec:data_sources}
\method{} is built on xBD, which provides pre- and post-disaster optical imagery, building footprints, disaster metadata, and ordinal damage labels.
We treat each annotated footprint as a building instance, with its pre/post imagery, disaster type, spatial location, and damage label, when available.
From these instances, we derive samples for single-building assessment, scene-level aggregation and reporting, and multi-instance comparison, selection, spatial reasoning, and prioritization.
Because xBD does not provide building-function labels, we enrich the instances with OSM-derived semantic information\cite{haklay2008openstreetmap}.

\subsection{OSM-Based Building-Function Enrichment}
\label{sec:benchmark_osm_enrichment}
xBD provides building footprints and post-disaster damage annotations, but no building-function labels. To enable function-aware disaster evaluation, 
we enrich xBD instances with OpenStreetMap (OSM) semantics, treating OSM as high-precision but incomplete supervision due to sparse and uneven coverage.
For each image, we query the Overpass API for nearby OSM building elements within a padded bounding box around all xBD footprints and match each footprint to candidates geometrically.

\begin{figure}[t]
\centering
\includegraphics[width=0.9\linewidth]{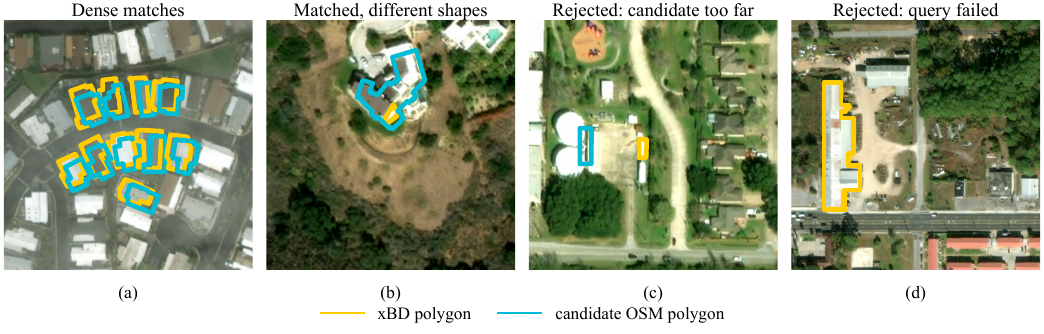}
\caption{OSM matching examples with accepted aligned footprints and rejected distant or missing matches. © OpenStreetMap contributors.}
\label{fig:osm_match_examples}
\end{figure}

When polygon geometries are available, we prioritize asymmetric xBD footprint coverage:
\begin{equation}
\mathrm{coverage}_{\mathrm{xBD}}(P_{\mathrm{xBD}}, P_{\mathrm{OSM}}) =
\frac{\mathrm{area}(P_{\mathrm{xBD}} \cap P_{\mathrm{OSM}})}
{\mathrm{area}(P_{\mathrm{xBD}})}.
\label{eq:osm_coverage}
\end{equation}
% This measures the fraction of the xBD footprint covered by the OSM polygon, rather than symmetric intersection-over-union.
% Matches above a predefined coverage threshold are accepted; otherwise, we fall back to center-point distance; candidates failing both criteria are rejected.
% We store confidence metadata including coverage ratio, nearest-candidate distance, number of candidates, acceptance mode, and query status.
This score measures the fraction of the xBD footprint covered by an OSM polygon, rather than the symmetric intersection-over-union.
For each xBD instance, we accept the candidate with the highest coverage when the score is at least 0.10. If no candidate satisfies this threshold, we accept the nearest candidate only when the haversine distance between representative geometry centers is at most 10,m; otherwise, the candidate is rejected.
Matching confidence is not a learned probability, but stored geometric metadata including coverage ratio, center-point distance, number of candidates, acceptance mode, and query status. The thresholds were chosen to tolerate modest xBD--OSM alignment differences while limiting matches to nearby candidates.
OSM annotations were obtained from the database state available at retrieval time, so matching enforces spatial correspondence but not event-specific historical alignment.
Figure~\ref{fig:osm_match_examples} shows representative accepted and rejected matches.

% After matching, we extract semantic tags from accepted OSM elements and select a primary tag using a fixed priority order that favors function-bearing keys such as amenity, shop, and office over generic keys such as building.
% We discard generic, ambiguous, or out-of-taxonomy tags, including building=yes, unknown, road/path, structural-only, agricultural, and vague place labels.
% The remaining tags are normalized into eight coarse building-function categories: residential, commercial, educational facilities, medical facilities, government/emergency, industrial/utilities, religious/cultural, and recreational/leisure.
After matching, we extract semantic tags from accepted OSM elements and select a primary tag using a fixed priority order that favors function-bearing keys such as amenity, shop, and office. We discard generic, ambiguous, or out-of-taxonomy tags, including building=yes, unknown, road/path, structural-only, agricultural, and vague place labels. The remaining tags are mapped to eight coarse building-function categories: residential, commercial, educational, medical, government/emergency, industrial/utilities, religious/cultural, and recreational/leisure.

The resulting subset is intentionally conservative: only buildings with sufficient geometric confidence and informative semantic tags are retained, so accepted OSM matches are distinct from final function-tagged buildings.
Across splits, approximately 48--55\% of xBD buildings receive an accepted OSM match, but only 12--15\% remain after semantic filtering.
The retained subset corresponds to 14.5\% of the training split and 14.4\% of the test split, reflecting real-world OSM incompleteness and yielding a long-tailed function-recognition benchmark.

\subsection{Task Taxonomy}
\label{subsec:task_taxonomy_statistics}
\begin{figure}[t]
\centering
\includegraphics[width=\textwidth]{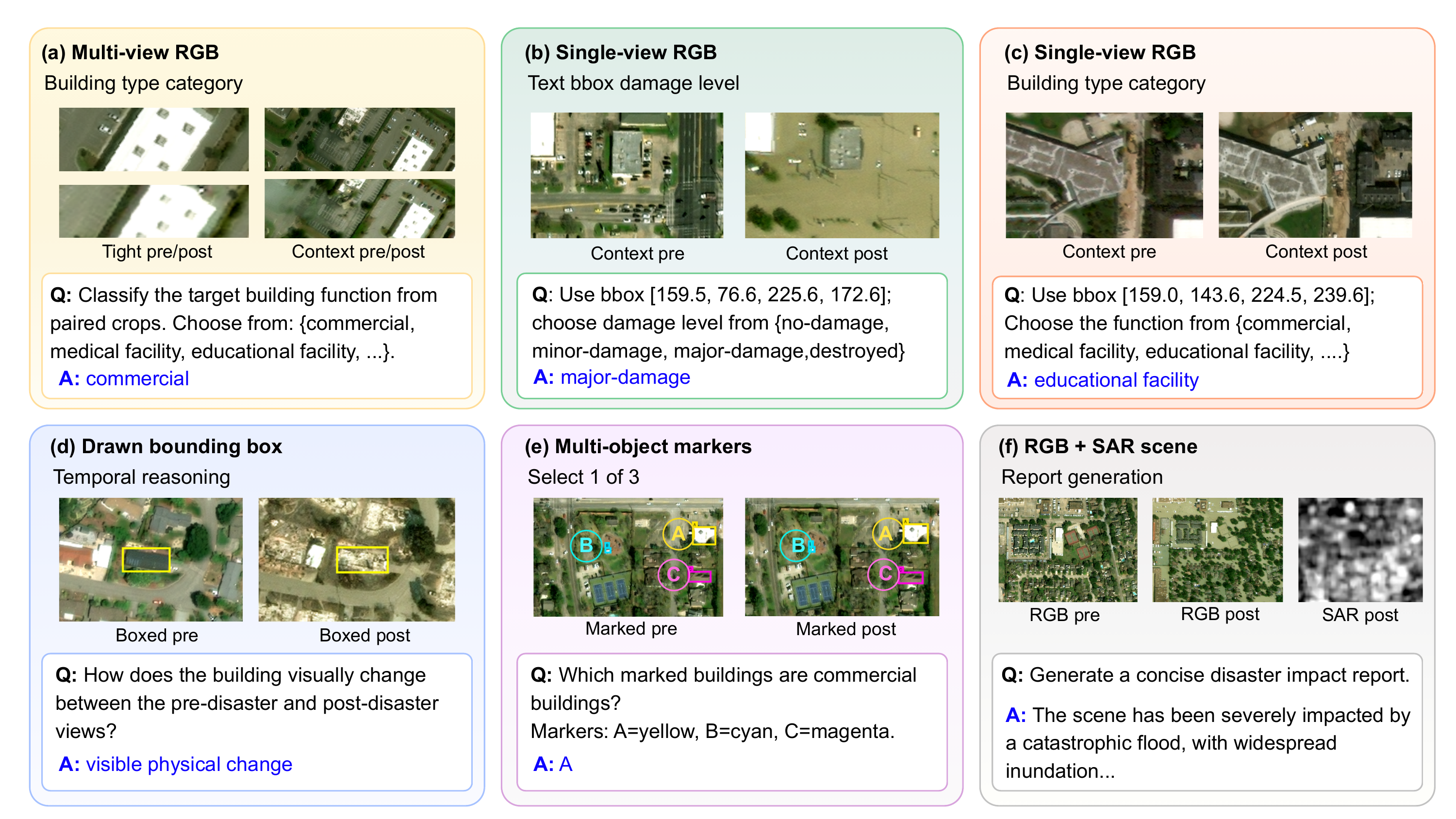}
\caption{Examples of \method{} task conversations.}
\label{fig:task_examples}
\end{figure}

\method{} contains 15 instruction-following task types grouped into instance-level, scene-level, report-generation, and multi-instance reasoning families, as summarized in Table~\ref{tab:task_taxonomy}.
Figure~\ref{fig:task_examples} shows representative task conversations across these input and task settings.

\begin{table}[t]
\centering
\footnotesize
\caption{DisasterInsight task taxonomy. The benchmark contains 15 task types grouped by evaluation family.}
\label{tab:task_taxonomy}
\renewcommand{\arraystretch}{1.22}
\setlength{\tabcolsep}{3.5pt}
\resizebox{\textwidth}{!}{%
\begin{tabular}{p{1.7cm} p{6.2cm} p{4.2cm} p{3.2cm}}
\toprule
\textbf{Family} & \textbf{Task types} & \textbf{Input} & \textbf{Metrics} \\
\midrule
Instance
& Building-function classification; damage-level estimation; temporal-change detection; joint function--damage assessment
& Target building views with pre/post imagery
& Acc., macro-F1, MAE, exact match \\
\midrule
Scene
& Total building count; damage-level count; function count
& Full pre/post scene imagery
& MAE, RMSE, key-wise MAE \\
\midrule
Report
& Short structured report generation; long structured report generation
& Full scene imagery; RGB--SAR scene setting
& BLEU-4, ROUGE-L, BERTScore \\
\midrule
Multi-instance
& Selection by damage; selection by function; joint function--damage selection; damage comparison; spatial reasoning; response-priority reasoning
& Marked buildings in pre/post scene imagery
& Acc., exact match, Set-F1 \\
\bottomrule
\end{tabular}%
}
\end{table}

The four task families are designed to test progressively richer forms of building-centric disaster understanding.
Instance-level tasks evaluate whether a model can identify the function, damage state, and temporal change of a specified building.
Scene-level tasks test whether building-level evidence can be aggregated into counts over the whole image.
Report-generation tasks evaluate whether models can communicate this evidence in a structured, response-oriented format.
Multi-instance tasks require models to compare or select among several marked buildings, testing whether they can combine visual damage cues, semantic building function, spatial relations, and response priority.

To formalize these task families, we represent each scene as 
$S=\{I^{\mathrm{pre}}, I^{\mathrm{post}}, \mathcal{B}\}$, where $I^{\mathrm{pre}}$ and $I^{\mathrm{post}}$ denote pre- and post-disaster imagery, and $\mathcal{B}=\{b_i\}_{i=1}^{N}$ is the set of annotated building instances.
Each building $b_i$ has a footprint or bounding box, a damage label $d_i \in \mathcal{D}$, and, when available, an OSM-derived function label $f_i \in \mathcal{F}$.
Instance-level tasks predict labels for a queried building $b_i$, scene-level and report-generation tasks predict aggregate or textual outputs over $\mathcal{B}$, and multi-instance tasks predict subsets or rankings over a marked set $\mathcal{M}\subset\mathcal{B}$.

\subsection{Benchmark Statistics}
\begin{figure}[t]
\centering
\includegraphics[width=0.90\textwidth]{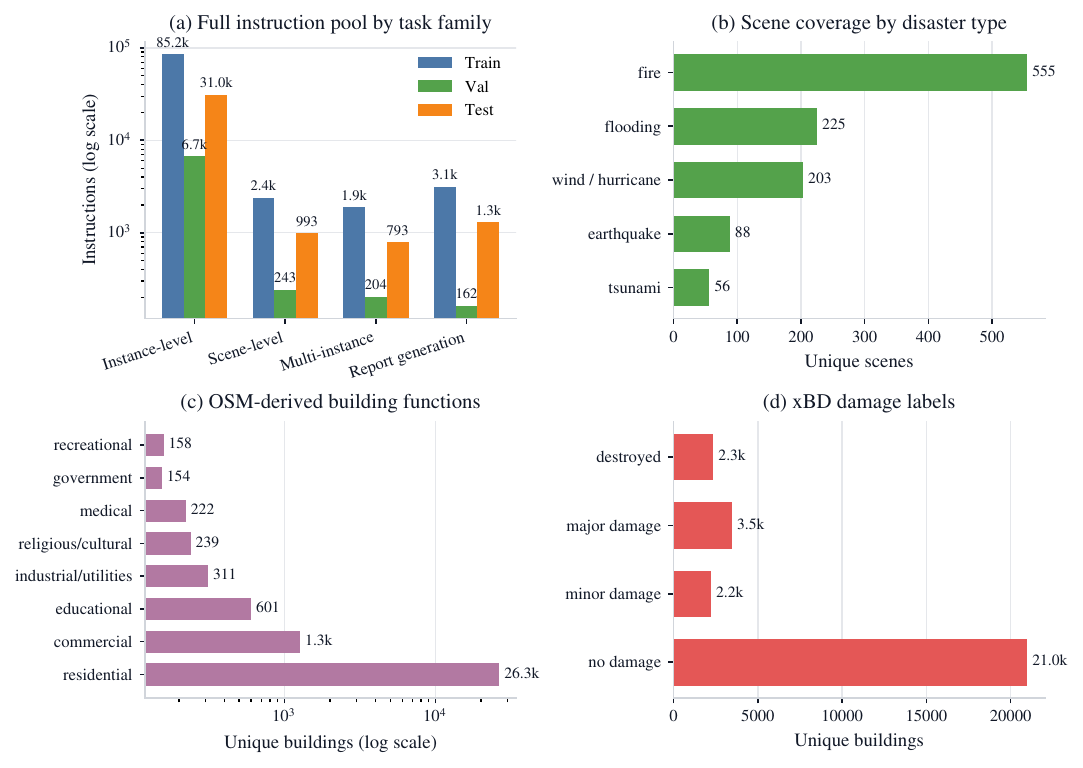}
\caption{
Benchmark statistics across task families, disaster types, building-function labels, and damage categories.
}
\label{fig:benchmark_statistics}
\end{figure}

Figure~\ref{fig:benchmark_statistics} summarizes the benchmark scale and label distributions.
The full instruction pool contains 134{,}108 task-specific records, with instance-level tasks forming the largest group.
The scenes cover fire, flooding, earthquake, tsunami, and wind-related hurricane events. Both OSM-derived building-function labels and xBD damage labels are imbalanced, reflecting realistic disaster-response data conditions.

\subsection{Input Settings and Grounding}
\label{sec:benchmark_multiview}
\method{} defines input configurations that vary in target grounding, spatial context, temporal evidence, and sensing modality.
For instance-level tasks, the target building is visually grounded in pre- and post-disaster crops, while scene-level tasks use full pre- and post-disaster images. Temporal tasks use both pre- and post-event imagery to test whether models can distinguish persistent building properties from disaster-induced changes.

For instance-level inputs, each target building can be represented at three spatial scales: \emph{tight}, \emph{context}, and \emph{global}. These views provide local appearance, surrounding context, and broader scene information, enabling diagnostic analysis of whether additional context helps or confuses building-centric reasoning.
We also define a single-view context variant, where instance-level examples use only the pre/post context crops.

For scene-level multimodal diagnostics, we provide co-registered Sentinel-1 SAR visualizations aligned with the xBD RGB pre/post imagery. The RGB/SAR scene-level records pair optical imagery with post-disaster SAR evidence, enabling zero-shot analysis of whether VLMs can exploit radar backscatter for disaster understanding.
\subsection{Structured Report-Generation Protocol}
\label{subsec:report_generation}

For report generation, we construct reference reports from compact structured scene summaries rather than raw building-level annotations. Each summary aggregates report-relevant evidence, including disaster type, total building footprints, damage-label availability, damage counts, affected and destroyed counts, known building-function counts, and coarse spatial damage patterns when available. We explicitly distinguish all detected buildings from buildings with known damage or function labels to avoid treating missing annotations as no damage or generalizing them to all buildings.

Reference reports are produced offline using a refined prompting protocol that emphasizes visually grounded and summary-supported attributes, such as damage magnitude, severity, affected structures, spatial patterns, building-function information, and uncertainty. The prompt discourages unsupported claims about locations, casualties, financial losses, infrastructure failures, or structural integrity, since these require external or unverifiable information. We generate both concise short reports and structured long reports covering situation overview, spatial damage pattern, building type information, safety considerations, and response priorities.

Human-in-the-loop review was used to select the report-generation model and refine the prompt. Llama~3.3-70B-Instruct~\cite{meta2024llama33} produced the most grounded and stable outputs among the tested instruction-tuned LLMs, so we use it to generate the final references. At evaluation time, VLMs generate reports directly from imagery, without access to the structured summaries, and are compared with the references using automatic text metrics and qualitative inspection for factual grounding, hallucination, and missing-data handling. Thus, report generation is used as a diagnostic benchmark for grounded building-level disaster reasoning rather than as an operational reporting system.

\subsection{Models and Training}
\label{subsec:models_training}
We evaluate LLaVA-OneVision~\cite{li2024llavaonevision},
Qwen2.5-VL-7B~\cite{bai2025qwen25vltechnicalreport},
Qwen3-VL-8B and Qwen3-VL-30B~\cite{bai2025qwen3},
and TEOChat~\cite{irvin2025teochat}, covering both general-purpose and remote-sensing-oriented VLMs. All models are evaluated on the same held-out standardized \method{} test set with identical prompts and image inputs for each task setting.

To reduce sensitivity to wording, each task type uses multiple prompt templates, and one template is sampled uniformly for each instance. The templates vary in tone and framing, including direct classification, analyst-style, conversational, and response-oriented disaster-assessment prompts, while preserving the same task definition and ground-truth target.

We release the full task-specific instruction pool, but use a standardized balanced split for experiments to avoid domination by high-volume task families. The split contains 7{,}563 training, 809 validation, and 3{,}171 test conversations across 15 task types, with train/validation/test conversations sampled only from their corresponding task-specific pools. Thus, no task-specific test examples are used during instruction tuning.

We instruction-tune Qwen3-VL-8B and TEOChat with LoRA for one epoch using learning rate $2\times10^{-5}$, dropout $0.05$, cosine scheduling, and warmup ratio $0.03$. Qwen3-VL-8B applies LoRA to attention and MLP projections, while TEOChat uses 8-bit LoRA with a frozen backbone and trainable multimodal projector. These settings were selected based on validation-set performance, training stability, and efficient adaptation across heterogeneous task types.

\subsection{Metrics and Evaluation Protocol}
\label{subsec:metrics_protocol}

We use task-specific metrics according to output type.
For categorical instance-level and multi-instance tasks, we report accuracy and F1 scores, emphasizing macro-F1 for long-tailed building-function labels. For damage-level classification, which follows an ordinal severity structure, we additionally report ordinal mean absolute error (MAE), so that predictions farther from the ground-truth severity level receive a larger penalty~\cite{edstedt2022vidharm}. For counting tasks, we report MAE and RMSE, with key-wise MAE for count-dictionary outputs. For multi-instance selection tasks, we report exact match and set-F1 to account for partially correct selected building sets.

For report generation, we compare model outputs with the offline reference reports using BLEU-4~\cite{papineni2002bleu}, ROUGE-L F1~\cite{lin2004rouge}, and BERTScore F1~\cite{zhang2020bertscore}. These metrics capture complementary aspects of text similarity, including n-gram precision, longest-common-subsequence overlap, and contextual semantic similarity. Because automatic text metrics do not fully capture factual grounding, we also inspect representative outputs for hallucination, unsupported claims, and incorrect handling of unavailable damage or function labels.

\section{Results}
\label{sec:results}
\subsection{Overall Benchmark Performance}
\label{subsec:overall_results}
Tables~\ref{tab:object_multi_results} and~\ref{tab:scene_family_results} report results on the standardized \method{} test split.
Table~\ref{tab:object_multi_results} covers instance-level tasks and representative multi-instance tasks, while Table~\ref{tab:scene_family_results} reports scene-level counting and report generation.
Instruction tuning improves most metrics, especially damage estimation, temporal-change recognition, scene-level counting, multi-instance reasoning, and report generation.
Qwen3-VL-8B + LoRA achieves the best building-function macro-F1, function-selection set-F1, spatial-reasoning accuracy, and report-generation scores, while TEOChat + LoRA obtains the best damage MAE and temporal-change scores.

For instance-level tasks, tuning improves damage-level estimation and temporal-change detection more strongly than building-function recognition.
For scene-level tasks, tuning reduces total-count error and substantially improves report metrics.
The remaining errors are analyzed in Section~\ref{sec:discussion}.
\begin{table*}[!t]
\centering
\small
\caption{
Instance- and multi-instance results on the standardized \method{} test split.
Func., Dmg., and Temp. denote building function, damage level, and temporal change.
F-Sel., Spatial, and Priority denote multi-instance function selection, spatial reasoning, and response-priority reasoning.
All values are percentages except Dmg. MAE.
}
\setlength{\tabcolsep}{3.5pt}
\renewcommand{\arraystretch}{1.12}
\resizebox{\textwidth}{!}{%
\begin{tabular}{lccccccccc}
\toprule
& \multicolumn{6}{c}{Instance-level}
& \multicolumn{3}{c}{Multi-instance} \\
\cmidrule(lr){2-7}
\cmidrule(lr){8-10}
& \multicolumn{2}{c}{Func.}
& \multicolumn{2}{c}{Dmg.}
& \multicolumn{2}{c}{Temp.}
& F-Sel.
& Spatial
& Priority \\
\cmidrule(lr){2-3}
\cmidrule(lr){4-5}
\cmidrule(lr){6-7}
\cmidrule(lr){8-10}
Model
& Acc. $\uparrow$ & mF1 $\uparrow$
& Acc. $\uparrow$ & MAE $\downarrow$
& Acc. $\uparrow$ & mF1 $\uparrow$
& Set-F1 $\uparrow$
& Acc. $\uparrow$
& Acc. $\uparrow$ \\
\midrule
\multicolumn{10}{l}{\textit{Open-source models}} \\
Qwen2.5-VL-7B
& 74.9 & 21.0
&  \textbf{81.2} & \textbf{0.32}
& 44.8 & 16.8
& 66.5
& 47.4
& \textbf{33.3} \\
Qwen3-VL-8B
& 86.6 & 25.9
& 72.7 & 0.46
& 44.1 & 27.9
& 66.4
& \textbf{53.4}
& 31.0 \\
Qwen3-VL-30B
& \textbf{91.0} & \textbf{30.1}
& 64.2 & 0.69
& 22.7 & 13.3
& \textbf{70.3}
& 39.8
& 31.0 \\
LLaVA-OneVision
& 85.6 & 24.0
& 77.1 & 0.38
& \textbf{61.5} & \textbf{28.9}
& 59.2
& 34.6
& \textbf{33.3} \\
\midrule
\multicolumn{10}{l}{\textit{Remote-sensing model}} \\
TEOChat
& 84.6 & 23.7
& 80.1 & \textbf{0.32}
& 31.8 & 26.5
& 66.5
& 33.8
& 14.3 \\
\midrule
\multicolumn{10}{l}{\textit{Fine-tuned models}} \\
Qwen3-VL-8B + LoRA
& \textbf{93.3} & \textbf{37.0}
& 88.9 & 0.16
& 89.5 & 52.1
& \textbf{82.8}
& \textbf{70.7}
& \textbf{38.1} \\
TEOChat + LoRA
& \textbf{93.3} & 24.0
& \textbf{90.0} & \textbf{0.14}
& \textbf{90.6} & \textbf{53.4}
& 73.2
& 65.4
& 28.6 \\
\bottomrule
\end{tabular}%
}
\vspace{2pt}
\label{tab:object_multi_results}
\end{table*}

\begin{table}[tb]
\centering
\small
\caption{
Scene-level counting and report-generation results.
Counting uses total-count MAE and key-wise MAE for function and damage counts; report metrics are BLEU-4, ROUGE-L F1, and BERTScore F1 percentages.
}
\setlength{\tabcolsep}{9.5pt}
\renewcommand{\arraystretch}{1}
\resizebox{0.93\textwidth}{!}{%
\begin{tabular}{lccc|ccc}
\toprule
& \multicolumn{3}{c|}{Counting}
& \multicolumn{3}{c}{Reports} \\
\cmidrule(lr){2-4}
\cmidrule(lr){5-7}
Model
& Tot. $\downarrow$
& Func. $\downarrow$
& Dmg. $\downarrow$
& B4 $\uparrow$
& R-L $\uparrow$
& BS $\uparrow$ \\
\midrule
\multicolumn{7}{l}{\textit{Open-source models}} \\
Qwen2.5-VL-7B
& \textbf{66.0} & 6.5 & 19.2 & 1.8 & 16.3 & 84.3 \\
Qwen3-VL-8B
& 175.6 & 9.8 & 20.5 & 3.1 & 17.7 & 84.8 \\
Qwen3-VL-30B
& 69.7 & 15.2 & \textbf{18.3} & 2.2 & 16.6 & 84.2 \\
LLaVA-OneVision
& 78.6 & \textbf{5.7} & 20.2 & 3.4 & 19.4 & 85.4 \\
\midrule
\multicolumn{7}{l}{\textit{Remote-sensing model}} \\
TEOChat
& 248.8 & 28.2 & 26.5 & \textbf{4.7} & \textbf{21.5} & \textbf{86.9} \\
\midrule
\multicolumn{7}{l}{\textit{Fine-tuned models}} \\
Qwen3-VL-8B + LoRA
& \textbf{25.2} & \textbf{2.7} & \textbf{10.2}
& \textbf{19.5} & \textbf{41.1} & \textbf{90.9} \\
TEOChat + LoRA
& 36.4 & 7.3 & 17.2
& 12.9 & 29.5 & 89.2 \\
\bottomrule
\end{tabular}%
}
\vspace{2pt}
\label{tab:scene_family_results}
\end{table}

\subsection{Qualitative Results}
\label{subsec:qualitative_results}

\begin{figure}[t]
    \centering
    \includegraphics[width=\linewidth]{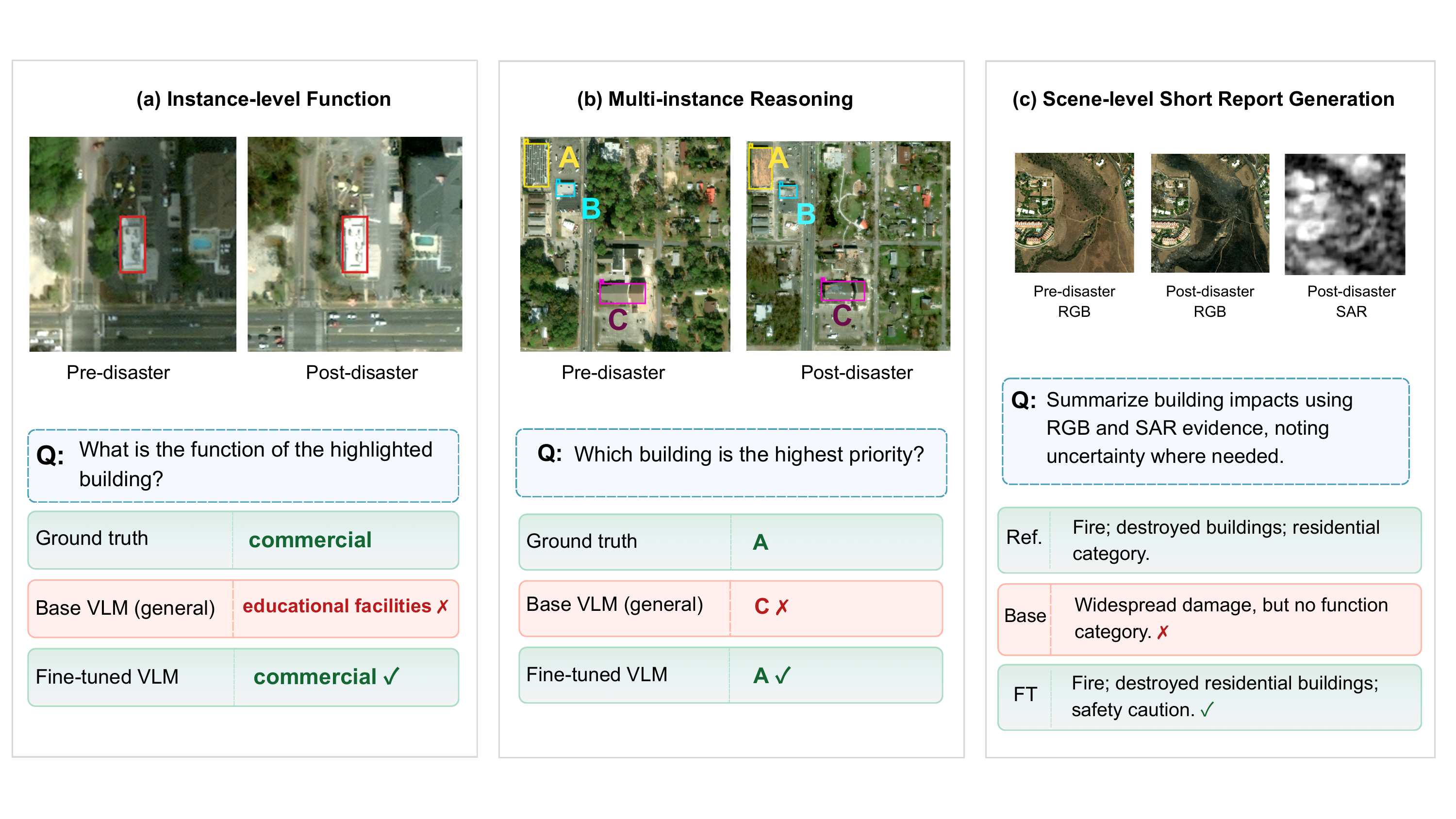}
    \caption{
    Qualitative examples across task families.
    }
    \label{fig:qualitative_results}
\end{figure}

Figure~\ref{fig:qualitative_results} shows representative predictions across instance-level, multi-instance, and scene-level report-generation tasks.
The examples illustrate that instruction tuning can improve grounding for building-function recognition, multi-instance reasoning, and report generation.
They also highlight the types of reasoning required by \method{}, including single-building assessment, scene-level summarization, and multi-building prioritization.
More detailed error patterns, including long-tailed building-function failures and response-priority mistakes, are discussed in Section~\ref{sec:discussion}. Additional task examples, OSM-enrichment details, and report-generation prompt details are provided in the supplementary material.
\section{Analysis and Discussion}
\label{sec:discussion}

\begin{figure}[t]
\centering
\includegraphics[width=0.9\linewidth]{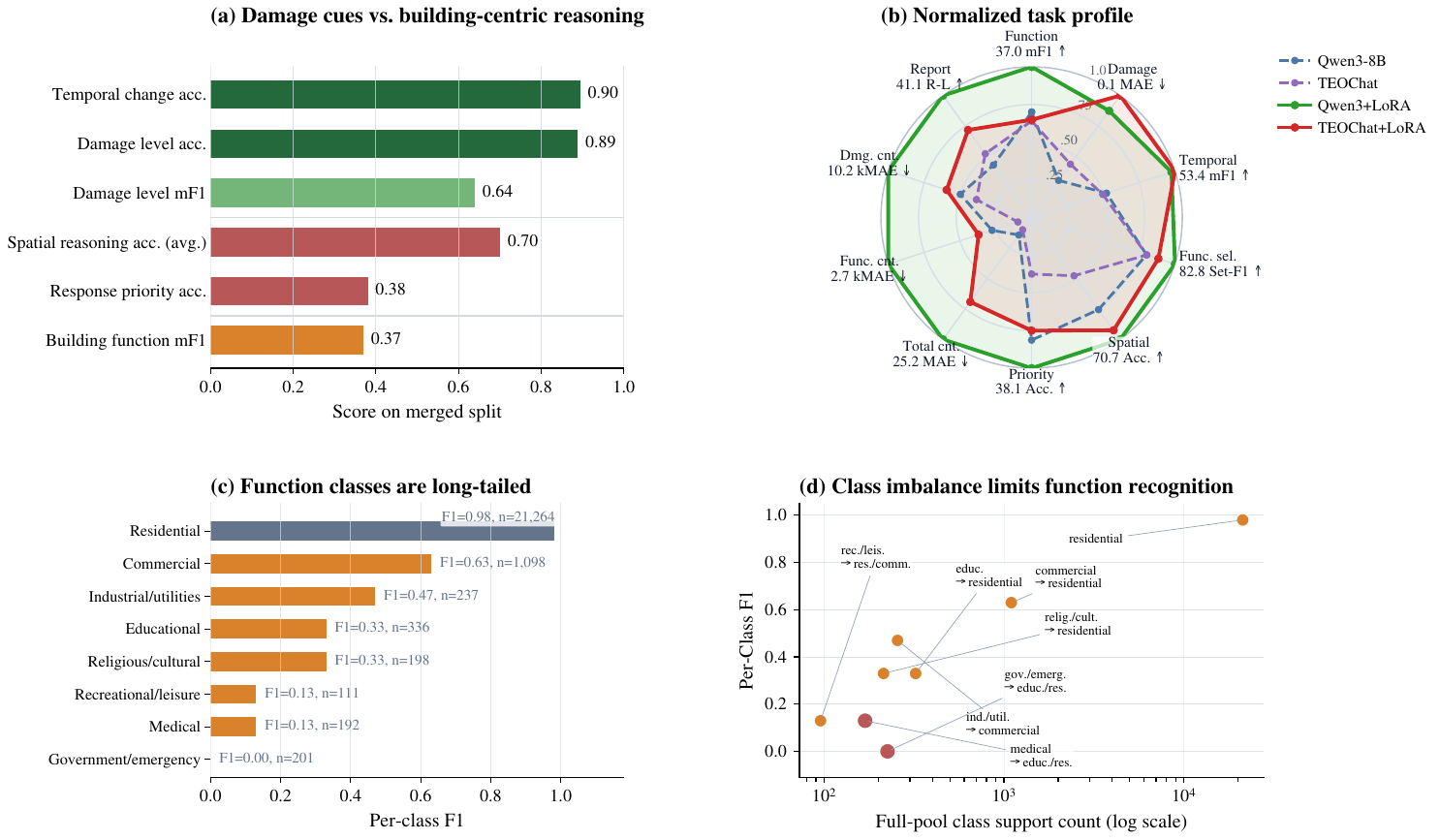}
\caption{
Gap-analysis summary.
}
\label{fig:analysis_chart}
\end{figure}

\textbf{Building-centric gap.} Figure~\ref{fig:analysis_chart} summarizes the main failure pattern revealed by \method{}.
Models are strongest on visually grounded damage cues: after instruction tuning, temporal-change detection, and damage-level estimation reach relatively high scores.
In contrast, performance drops for building-centric semantics and response-oriented reasoning.
Building-function recognition remains limited under macro-F1, and response-priority reasoning is among the weakest multi-instance tasks, showing that models struggle when decisions require combining visible damage evidence with the functional role of the affected building.

\textbf{Tuning effect.} The normalized task profile in Figure~\ref{fig:analysis_chart} (b) further shows that instruction tuning improves most task families, especially damage estimation, temporal change, counting, and report generation.
However, the improvement is uneven.
The tuned Qwen3-VL model performs strongly on several axes, but the building-function axis remains far from saturated.
This suggests that instruction tuning improves benchmark alignment and visible-cue reasoning, but does not fully solve semantic building understanding.

\textbf{Long-tail limitation.} One reason for this semantic gap is the long-tailed OSM-derived function distribution shown in Figure~\ref{fig:analysis_chart} (c).
Residential buildings dominate the full instruction pool, while response-critical classes such as medical, government/emergency, and recreational/leisure facilities have much lower support.
In Figure~\ref{fig:analysis_chart} (d), per-class F1 follows this imbalance: residential buildings achieve high recognition performance, whereas rare functional categories remain brittle and are often confused with more common classes.
This matters for disaster response because rare categories can carry high operational priority despite having limited training support.

\textbf{Input formulation.} Changing the input formulation does not by itself resolve the building-centric gap.
In auxiliary view diagnostics, additional context can help some damage-related predictions, but it does not consistently improve building-function recognition.
Similarly, SAR-aware scene prompts provide a useful benchmark extension for future multimodal analysis, but our main results do not show that SAR prompting alone closes the gap.
We therefore treat view design and SAR/RGB inputs as diagnostic extensions rather than solutions to the core semantic reasoning problem.

Overall, these results support our central claim: current VLMs are more reliable at recognizing visible damage cues than at grounding higher-level semantic reasoning in building-level visual evidence.
This suggests that part of the bottleneck lies in connecting visual evidence to semantic building information during downstream disaster-response reasoning.

\FloatBarrier
\section{Conclusion}
\label{sec:conclusion}

We introduced \method{}, an instance-centric benchmark for evaluating vision--language models in disaster-response remote sensing.
By enriching {xBD} building instances with OSM-derived functional labels and constructing instruction-following tasks across instance-level assessment, scene-level counting, multi-instance reasoning, and structured report generation, \method{} enables diagnostic evaluation beyond visible damage cues.

Our experiments show a consistent building-centric gap.
Instruction tuning improves damage estimation, temporal-change reasoning, counting, and report-generation metrics, but models remain limited in building-function recognition, response-priority reasoning, and faithful use of building-level evidence in reports.
These limitations are amplified by long-tailed building functions, especially rare but operationally important categories such as medical, government/emergency, and recreational/leisure facilities.
Auxiliary input diagnostics also suggest that additional views or SAR/RGB inputs are not sufficient by themselves to solve semantic building understanding.

These findings indicate that current VLMs are more reliable at recognizing visible damage cues than at grounding higher-level semantic reasoning in building-level visual evidence.
\method{} provides a controlled benchmark for studying these limitations and motivates future work on long-tail-aware learning, stronger geospatial grounding, multimodal adaptation, and uncertainty-aware disaster-response VLMs.

\section*{Acknowledgements}
This work was supported in part by the Excellence Center at Linköping-Lund in Information Technology (ELLIIT) Researcher Funding, the Zenith Research Program, and the Swedish Research Council with grant agreement no. 2022-04266 and no. 2024-05652. The computational resources were provided by the National Academic Infrastructure for Supercomputing in Sweden (NAISS) at C3SE, partially funded by the Swedish Research Council through grant agreement no. 2022-06725, and by the Berzelius resource, provided by the Knut and Alice Wallenberg Foundation at the National Supercomputer Centre.

\bibliographystyle{splncs04}
\bibliography{refs}

\clearpage
\section*{Appendix}
\appendix
\setcounter{table}{0}
\setcounter{figure}{0}
\renewcommand{\thetable}{A.\arabic{table}}
\renewcommand{\thefigure}{A.\arabic{figure}}

\subsection*{OSM Enrichment Details}
\label{sec}
\paragraph{Implementation details.}
As described in the main paper, we enrich xBD building footprints with OSM-derived semantic information using image-level OSM queries, geometric matching, and semantic filtering.
This appendix provides additional implementation details and representative tag mappings.
For each xBD scene, OSM elements are queried within the georeferenced image extent rather than separately for each building, reducing redundant requests and improving reproducibility.
Candidate OSM elements are matched to xBD footprints using polygon overlap when available, with centroid distance as a fallback.
For each attempted match, we store confidence metadata including overlap ratio, nearest-candidate distance, number of candidates, acceptance mode, and query status.

\paragraph{Semantic tag mapping.}
As shown in table \ref{tab:osm_category_mapping}, for accepted OSM matches, semantic tags are normalized into the eight coarse building-function categories used in the benchmark: residential, commercial, educational facilities, medical facilities, government/emergency, industrial/utilities, religious/cultural, and recreational/leisure.
When multiple tags are present, we prioritize function-bearing keys such as \texttt{amenity}, \texttt{shop}, and \texttt{office} over generic keys such as \texttt{building}.
Generic, ambiguous, or out-of-taxonomy tags such as \texttt{building=yes}, unknown labels, road/path labels, structural-only labels, agricultural labels, and vague place labels are discarded.

\begin{table}[t]
\centering
\caption{Representative mapping from normalized OSM semantic labels to benchmark building-use categories.}
\label{tab:osm_category_mapping}
\small
\begin{tabular}{p{0.62\linewidth} p{0.28\linewidth}}
\toprule
\textbf{OSM Labels} & \textbf{Benchmark Category} \\
\midrule
\texttt{house}, \texttt{apartments}, \texttt{residential}, \texttt{detached}, \texttt{bungalow}, \texttt{static\_caravan} & residential \\
\texttt{hospital}, \texttt{clinic}, \texttt{pharmacy}, \texttt{doctors}, \texttt{dentist}, \texttt{veterinary} & medical\_facilities \\
\texttt{school}, \texttt{college}, \texttt{university}, \texttt{kindergarten}, \texttt{childcare} & educational\_facilities \\
\texttt{fire\_station}, \texttt{police}, \texttt{courthouse}, \texttt{shelter}, \texttt{prison}, \texttt{townhall}, \texttt{government} & government\_emergency \\
\texttt{supermarket}, \texttt{convenience}, \texttt{mall}, \texttt{restaurant}, \texttt{fast\_food}, \texttt{bank}, \texttt{retail}, \texttt{commercial}, \texttt{doityourself}, \texttt{computer}, \texttt{car\_wash}, \texttt{pet} & commercial \\
\texttt{warehouse}, \texttt{factory}, \texttt{storage\_rental}, \texttt{industrial}, \texttt{fuel}, \texttt{power}, \texttt{works} & industrial\_utilities \\
\texttt{church}, \texttt{place\_of\_worship}, \texttt{mosque}, \texttt{temple}, \texttt{museum}, \texttt{cemetery} & religious\_cultural \\
\texttt{swimming\_pool}, \texttt{park}, \texttt{hotel}, \texttt{motel}, \texttt{sports\_centre}, \texttt{pitch}, \texttt{theatre} & recreational\_leisure \\
\bottomrule
\end{tabular}
\end{table}

\begin{figure}[t]
\centering
\begin{subfigure}{0.48\linewidth}
\centering
\includegraphics[width=\linewidth]{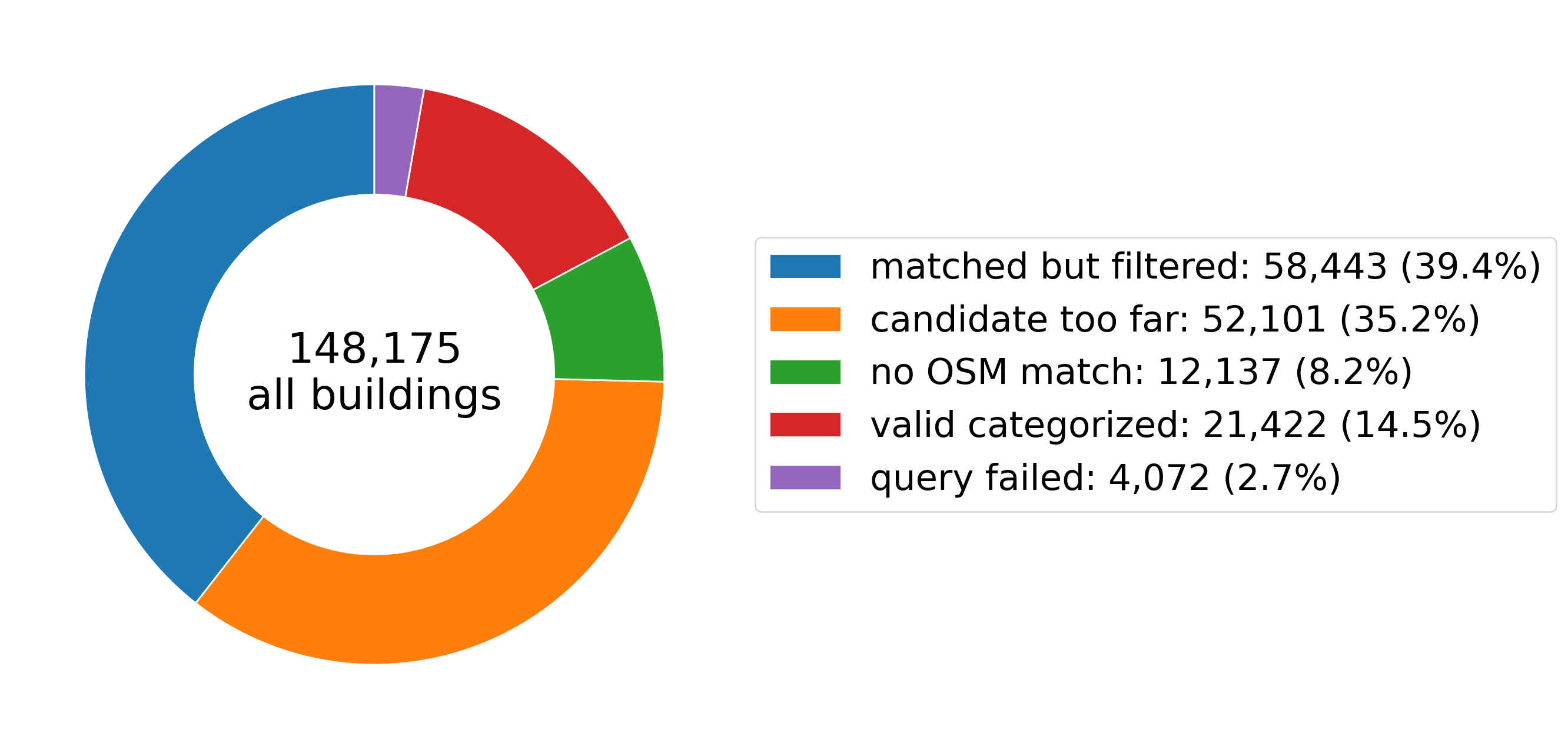}
\caption{Training split}
\end{subfigure}
\hfill
\begin{subfigure}{0.48\linewidth}
\centering
\includegraphics[width=\linewidth]{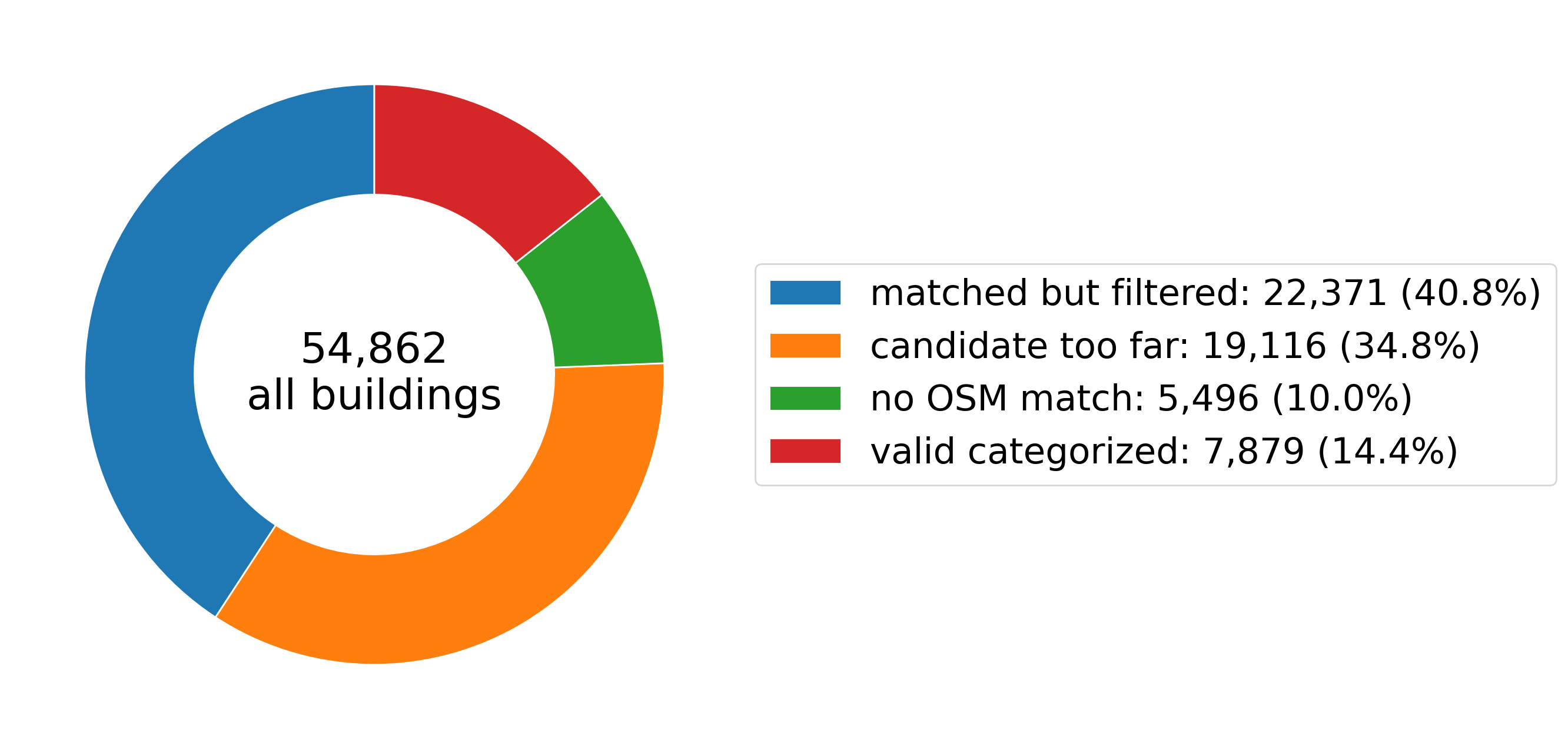}
\caption{Test split}
\end{subfigure}
\caption{
OSM enrichment filtering statistics for xBD buildings.
The final function-tagged subset is intentionally conservative, reflecting sparse and noisy OSM coverage.
}
\label{fig:osm_filtering}
\end{figure}

\paragraph{Coverage and filtering statistics.}
Figure~\ref{fig:osm_filtering} summarizes the filtering outcome for the xBD training and test splits.
In the training split, 79{,}865 of 148{,}175 buildings obtain an accepted OSM match, but only 21{,}422 are retained as valid function-tagged examples after semantic filtering.
In the test split, 30{,}250 of 54{,}862 buildings obtain an accepted OSM match, with 7{,}879 retained as valid function-tagged examples.
Thus, the final conservative functional subset corresponds to 14.5\% of training buildings and 14.4\% of test buildings.
The validation split used in the benchmark is sampled after filtering and is therefore not interpreted as a raw OSM coverage distribution.

\paragraph{Limitations.}
The OSM-derived labels are weak semantic annotations rather than exhaustive building-use labels.
They are sparse, geographically uneven, and long-tailed, with residential and commercial categories dominating the retained subset.
We therefore use them for diagnostic function-aware evaluation rather than as complete ground truth for all xBD buildings.
This conservative design improves label precision but makes rare response-critical categories, such as medical and government/emergency buildings, especially challenging.
\subsection*{Prompt Refinement for Report Generation}
\label{app:report_prompt_refinement}

\paragraph{Structured scene summaries.}
For report-generation references, we convert building-level annotations into compact structured scene summaries.
Each summary aggregates report-relevant evidence, including disaster type, total detected buildings, damage-label availability, damage-level counts, affected and destroyed counts, known building-function counts, and coarse spatial damage patterns when available.
The summary explicitly distinguishes all detected xBD buildings from buildings with known damage labels and buildings with known OSM-derived function labels, avoiding the interpretation of missing labels as no damage or as representative of all buildings.

A representative summary is:
\begin{quote}
\small
\ttfamily
\begin{minipage}{0.95\linewidth}
\raggedright
disaster=flooding; xbd\_buildings\_total=51;\\
damage\_scored\_buildings=45; damage\_unavailable\_buildings=6;\\
damage\_counts=no-damage:27, minor-damage:7, major-damage:7, destroyed:4;\\
affected\_total=18; destroyed\_total=4;\\
category\_known=3; category\_unknown=48;\\
known\_category\_counts=residential:2, commercial:1;\\
damage\_region\_order\_by\_severity=east$>$north$>$south-west;\\
top\_damage\_regions=east:8 affected,3 destroyed; north:5 affected,1 destroyed;\\
south-west:2 affected,0 destroyed
\end{minipage}
\end{quote}

\paragraph{Prompt refinement.}
We refined the report-generation prompt by manually inspecting sampled outputs during development.
Early variants sometimes produced fluent but weakly grounded reports, including unsupported event details, overestimated damage severity, generic recommendations, repetition in long reports, and ambiguous handling of missing labels.
We therefore revised the prompt to require the generator to use only the structured summary, avoid unsupported claims, preserve uncertainty when labels are unavailable, and keep known building-function information separate from total building counts.

\paragraph{Final reference-generation prompt.}
The final prompt constrains the offline report generator to produce grounded short and long references from the structured scene summary:

\begin{tcolorbox}[
colback=gray!3,
colframe=gray!45,
title=\textbf{Report-generation prompt template},
fonttitle=\small,
sharp corners,
boxrule=0.5pt
]
\small
\textbf{System prompt:} Generate a response-oriented disaster impact report from structured satellite-derived metadata. Use only the provided information and do not introduce external details.

\textbf{Key rules:}
\begin{itemize}
\item Total buildings and damage statistics must be interpreted separately from unavailable damage information.
\item Unavailable damage status must not be treated as damaged or undamaged.
\item Functional categories apply only to buildings with known category information.
\item Do not introduce unsupported details such as named events, locations, infrastructure failures, financial losses, casualties, or structural assessments.
\end{itemize}

\textbf{Conditional rules:}
\begin{itemize}
\item If no damage information is available, explicitly state this without implying absence of damage.
\item If no affected buildings are recorded, keep safety and response recommendations brief.
\item Only describe spatial damage patterns when supported by the input.
\end{itemize}

\textbf{Short report:} One concise paragraph (60--100 words).

\textbf{Long report:} Structured output with sections: Situation Overview, Spatial Damage Pattern, Building Type Information, Safety Considerations, and Response Priorities.
\end{tcolorbox}

\paragraph{Evaluation.}
At evaluation time, VLMs do not receive the structured summaries.
They generate reports directly from the image inputs, including SAR when present in the task setting, and their outputs are compared against the reference reports using BLEU-4, ROUGE-L, and BERTScore.
We treat these automatic metrics as approximate indicators of report similarity, since they do not fully capture factual grounding or hallucination.
Accordingly, report generation in \method{} is intended as a diagnostic benchmark task rather than an operational disaster-reporting system.
\subsection*{Additional Benchmark Examples}
\label{app:benchmark_examples}

Figure~\ref{fig} shows additional representative \method{} conversations across task families. These examples illustrate how visual inputs, natural-language instructions, and structured ground-truth answers are represented in the released benchmark. The examples cover instance-level joint building-function and damage assessment, multi-instance spatial reasoning, and multi-instance damage comparison. The benchmark is released in a tabular format suitable for instruction tuning and evaluation, with fields such as task type, split, messages, image references, and ground-truth annotations.

\begin{figure}[t]
\centering
\includegraphics[width=0.85\linewidth]{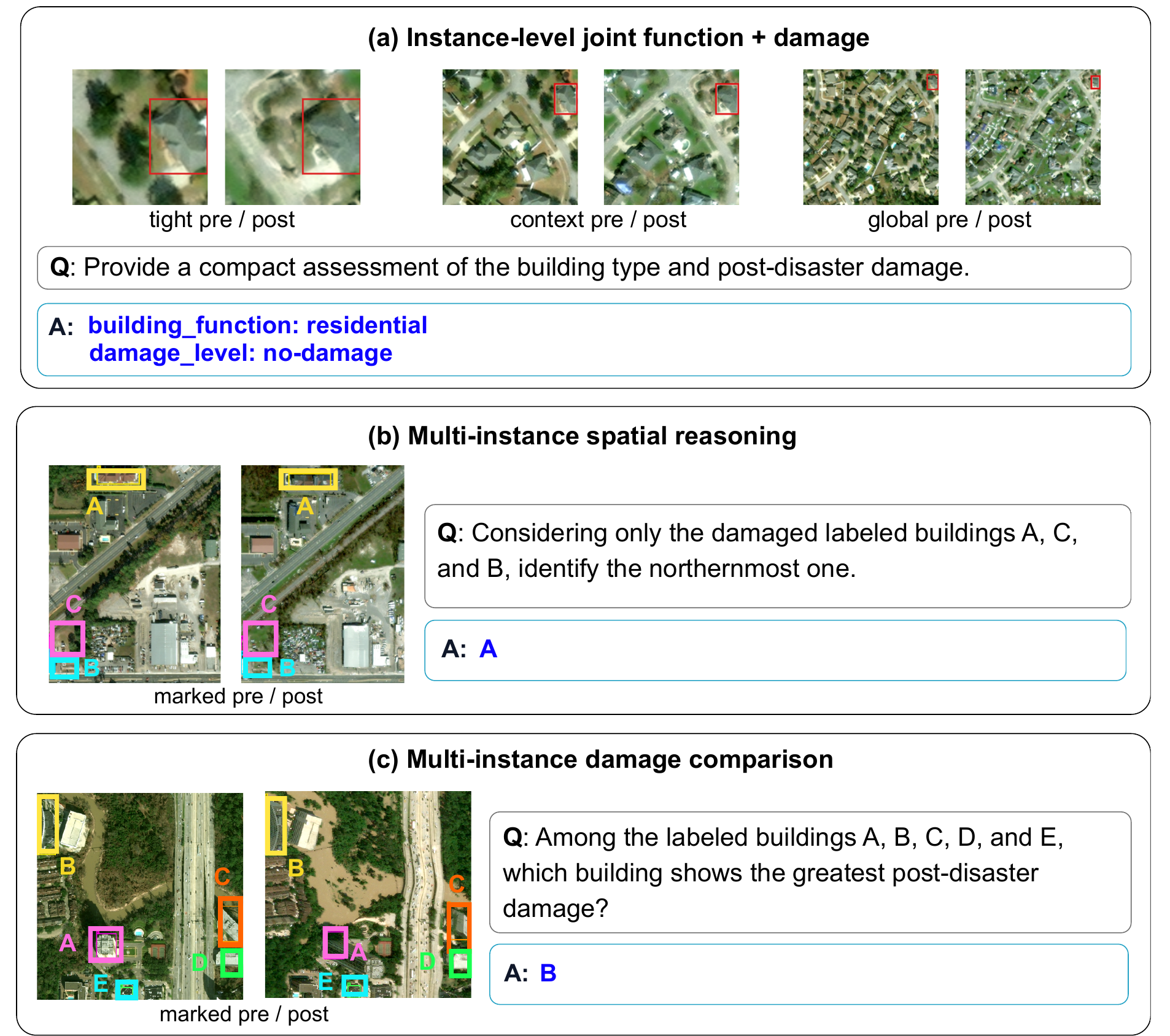}
\caption{Additional representative examples from \method{}.}

\label{fig}
\end{figure}

\end{document}